\documentclass[10pt,twocolumn,letterpaper]{article}

\usepackage[accsupp]{axessibility} 
\usepackage{graphicx}
\usepackage{float}
\usepackage{geometry}
\usepackage{amsmath} 
\usepackage{tabularx}
\usepackage{amssymb}
\usepackage{multirow}
\usepackage{bm}
\usepackage{listings}
\usepackage{xcolor}
\usepackage[pagenumbers]{cvpr} 

\usepackage{bm}
\usepackage{xcolor}

\newcommand\blfootnote[1]{%
	\begingroup
	\renewcommand\thefootnote{}\footnote{#1}%
	\addtocounter{footnote}{-1}%
	\endgroup
}

\definecolor{cvprblue}{rgb}{0.21,0.49,0.74}
\usepackage[pagebackref,breaklinks,colorlinks,citecolor=cvprblue]{hyperref}

\def\paperID{5658} 
\def\confName{CVPR}
\def\confYear{2024}

\title{Lodge: A Coarse to Fine Diffusion Network for \\ Long Dance Generation Guided by the Characteristic Dance Primitives}

\author{{Ronghui Li}$^{1,2}$, {YuXiang Zhang}$^{1}$,Yachao Zhang$^1$, \\Hongwen Zhang$^4$,Jie Guo$^{2}$,Yan Zhang$^{3\ddagger}$,Yebin Liu$^1$, Xiu Li$^{1\dagger}$\\$^1$Tsinghua University, $^2$Peng Cheng Laboratory $^3$Meshcapade, $^4$Beijing Normal University}

\definecolor{codegreen}{rgb}{0,0.6,0}
\definecolor{codegray}{rgb}{0.5,0.5,0.5}
\definecolor{codepurple}{rgb}{0.58,0,0.82}
\definecolor{backcolour}{rgb}{0.95,0.95,0.92}

\lstdefinestyle{mystyle}{
  backgroundcolor=\color{backcolour}, commentstyle=\color{codegreen},
  keywordstyle=\color{magenta},
  numberstyle=\tiny\color{codegray},
  stringstyle=\color{codepurple},
  basicstyle=\ttfamily\footnotesize,
  breakatwhitespace=false,         
  breaklines=true,                 
  captionpos=b,                    
  keepspaces=true,                 
  numbers=left,                    
  numbersep=5pt,                  
  showspaces=false,                
  showstringspaces=false,
  showtabs=false,                  
  tabsize=2
}

\begin{document}

\twocolumn[{ 
\maketitle
\begin{figure}[H]
\hsize=\textwidth  
\centering
    \includegraphics[width=\textwidth]{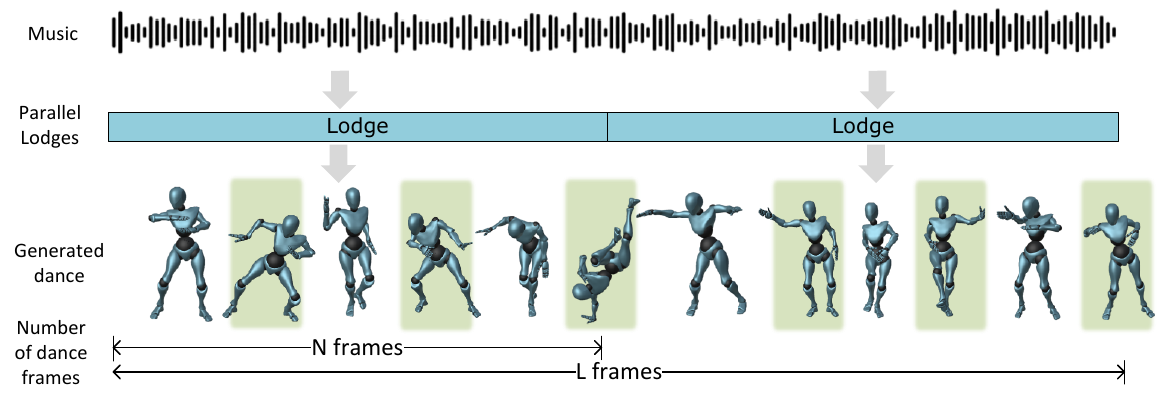}
    \caption{Lodge can parallelly generate extremely long dance. The sections highlighted in green represent the characteristic dance primitives. These are expressive 8-frame movements that not only support parallel generation but also contains choreographic patterns. They guide the diffusion network to generate long, expressive dances in parallel while adhering to choreographic rules.}
    \label{fig:framework}
\end{figure}
}]
\begin{abstract}
We propose Lodge, a network capable of generating extremely long dance sequences conditioned on given music. 
We design Lodge as a two-stage coarse to fine diffusion architecture, and propose the characteristic dance primitives that possess significant expressiveness as intermediate representations between two diffusion models. 
The first stage is global diffusion, which focuses on comprehending the coarse-level music-dance correlation and production characteristic dance primitives. In contrast, the second-stage is the local diffusion, which parallelly generates detailed motion sequences under the guidance of the dance primitives and choreographic rules.
\blfootnote{$\dagger$ corresponding author}
\blfootnote{$\ddagger$ This work was done while YZ was at ETH Z\"{u}rich.}
In addition, we propose a Foot Refine Block to optimize the contact between the feet and the ground, enhancing the physical realism of the motion.
Our approach can parallelly generate dance sequences of extremely long length, striking a balance between global choreographic patterns and local motion quality and expressiveness. Extensive experiments validate the efficacy of our method. Code, models, and demonstrative video results are available at: \href{https://li-ronghui.github.io/lodge}{https://li-ronghui.github.io/lodge}
\end{abstract}

\section{Introduction}
\label{sec:intro}

Given a piece of \textit{long-term} music, we aim at generating high-fidelity and diverse 3D dance motions in an automatic and efficient manner.
An effective solution is desired not only in many applications e.g. movie and game production, but also of high potential to inspire dance designers with novel movements, and improve their productivity.

With rapid advances in generative AI in recent years, existing methods~\cite{finedance,bailando,bailando++,kim2022brand,ofli2011learn2dance} demonstrated the ability to generate dance for seconds. However, dance in real applications often lasts for minutes. Dance performances and social dance usually last 3 to 5 minutes. Dance theater can last for more than 15 minutes or even an hour.
Therefore, the extremely long dance generation is becoming increasingly important as the demand for engaging dance content continues to grow.

However, generating long dance sequences poses a notable challenge due to the substantial computational resources needed for training.
Therefore, many methods are based on autoregressive models~\cite{aist++,dancere,bailando}, and continuously generate dance movements based on a relatively small sliding window. 
This autoregressive nature accumulates the model prediction errors as time progresses, and prevents the model from learning global choreographic patterns. As a result, motion freezing often occurs after several seconds~\cite{yang2023longdancediff}.
There are also some methods~\cite{bailando,gtn,zhang2021lightweight} maintain a latent space to represent motion, and combine a autoregressive based sequence model to learn music-dance paired relationship. However, the compressed latent space with limited representational capacity also makes these methods prone to overfitting, resulting in poor generalization and diversity.
Recently, EDGE~\cite{edge} proposed a diffusion-based dance generation model. 
During the denoising process, EDGE parallelly generate multiple dance segments with overlap while maintaining consistency between these overlapping parts using diffusion inpainting ~\cite{repaint}, and finally splices these segments into a long dance by linear interpolation.
However, their dances lack an overall choreographic structure and shows incoherence at the splices frames.

In summary, these existing methods regarding dance generation solely as a sequence-to-sequence problem. They struggle to enhance the dance quality of fine-grained local details while neglect the coarse-level global choreography patterns between music and dance. 
Referring to~\cite{dance_design,dance_design2,blom1982intimate,choreomaster}, dance is normally choreographed in a coarse-to-fine manner. 
Provided the entire music, dance designers first analyze the music attributes such as rhythm, genre and emotional tone, and create ``dance phrases'', i.e. some short-term expressive movements, which possess powerful expressiveness and richer semantic information.
During this stage,  dance designers can concentrate on  design characteristic dance phrases, such as ``inversions' and ``moonwalks''. 
Arrange these characteristic dance phrases follow the structured information of the music, the overall dance structure is laid down.
Subsequently, the entire dance is created by connecting dance phrases with transition movements.

Following the above insights, 
we think that the ``dance phrases'' contains  abundant distinctive movements and can convey global choreographic patterns. 
Therefore, similar to dance phrases, we propose \textit{characteristic dance primitives} suitable for network learning.
These dance primitives are expressive 8-frame key motions with high kinematic energy, with the following main advantages: (1) They are sparse, which reduce the computational demand. (2) They  have rich semantically information, and can transfer choreographic patterns. (3) They possess expressive motion characteristics, which can guide motion diffusion model to generate more dynamic movements and avoiding monotony.

Next, we design a coarse-to-fine dance generation framework with two motion diffusion models and employ the \textit{characteristic dance primitives} as their intermediate representation. 
The first stage is coarse-grained global diffusion, which takes as input long music and produces characteristic dance primitives. 
According to the fundamental choreographic rules, details in Sec. \ref{sec:method}, these dance primitives are further augmented to align with the beats and structural information of the music.
Subsequently, we employ parallel local diffusion to independently generate short dance segments. 
Based on some auto-selected dance primitives, we utilize diffusion guidance to strictly constrain consistency between the beginnings and ends of these segments. Therefore, these dance segments can be concatenated into a continuous long dance.
Simultaneously, under the guidance of the other dance primitives, the quality, expressiveness, and diversity of each dance segment are enhanced.


In addition, to improve the motion realism and eliminate foot-skating artifacts, we introduce a foot refine block inspired by ~\cite{xiang}.
We find it is difficult to simply use foot-related losses~\cite{lemo} to optimize the  SMPL~\cite{SMPL} format motion rotation data, especially in complex dance movements.
This is because the optimization objective exists in the linear joint position space while the SMPL format rotation data is mainly in nonlinear rotation space, and there is a domain gap hindering loss convergence.
Therefore, we compute foot contact information and utilize the foot refine block to generate modification values addressing foot skating.

In summary, our main contributions are as follows:
\begin{itemize}
    \item We introduce a coarse-to-fine diffusion framework that can produce long dances in a parallel manner. Our method is capable of learning the overall choreographic patterns while ensuring the quality of local movements.
    \item We propose the characteristic dance primitives that possess significant expressiveness as intermediate representations between two diffusion models.
    \item We propose a foot refine block and employ a foot-ground contact loss to eliminate artifacts such as skating, floating, and ground inter-penetration.
\end{itemize}

\section{Related Works}
\label{sec:Related}

\subsection{Human Motion Synthesis}
Human motion generation is an important task in the fields of computer vision and computer graphics.
Researchers make significant contributions in this direction. For instance, MDM \cite{MDM} successfully applies diffusion to the Text2Motion task, yielding high-quality motion results; GestureDiffuCLIP \citet{gesturediffuclip} achieves coordinated motion generation with speech and integrates style control through text and video guidance; SAGA\cite{SAGA} and Grasping\cite{Grasping} focuses on natural grasping motion generation;  \cite{DIMOS, zhang2022wanderings, huang2023diffusion} 
can produce human motions that interact with 3D scenes while avoiding collisions. CALM~\cite{CALM} and ASE~\cite{ASE} introduce reinforcement learning and physical simulation environments to enhance the physical realism of generated movements.
Despite substantial progress in aspects like motion quality, diversity, controllability, interactivity, and physical realism, etc, generating dance motions remains a challenging problem due to the inherent complexity and long-duration nature of dance movements.

\subsection{Music Driven Dance Generation}
Numerous studies aim to generate high-quality dance that synchronizes with the input music.
These approaches encompass various categories, including motion-graph methods~\cite{choreographers}, sequence model based methods~\cite{aist++, kim2022brand, bailando}, VQ-VAE based methods~\cite{bailando, gtn}, GAN-base methods~\cite{kim2022brand}, and diffusion based methods ~\cite{edge, finedance}.

The traditional motion-graph based methods \cite{ofli2011learn2dance, manfre2016automatic, berman2015kinetic} address this task as a similarity-based retrieval problem, which limits the diversity and creativity of generations.
In recent years, deep learning models have gained significant prominence, yielding aesthetically appealing outcomes. 
In sequence-based methods, LSTM \cite{hochreiter1997long} and Transformer~\cite{transformer} networks are commonly employed. These networks typically take as input music and the preceding dance sequence, predicting the subsequent dance in an autoregressive manner. 
Li et al. propose FACT\cite{aist++}, which inputs music and seed motions into a Transformer network, generating new dance frame by frame in an autoregressive manner, but challenges such as error accumulation and motion freezing~\cite{music2dance} phenomena persist. 
Based on VQ-VAE, Bailando incorporates a reinforcement learning-based action evaluator to optimize rhythm, while TM2D encodes the text-paired motion and music-paired dance into a shared codebook to achieve semantically controllable dance generation.
The advantages of VQ-VAE lie in its ability to maintain a pre-trained codebook, ensuring the motion quality of decoded dance sequences. But the codebook also limits dance diversity and hinders the network's generalization.
The Generative Adversarial Network (GAN) consists of a generator and a discriminator, engaged in adversarial training to produce realistic data. MNET \cite{kim2022brand} proposes a transformer-based dance generator and a multi-genre dance discriminator network to generate realistic dance clips and achieve genre control.
However, these GAN-based methods suffer from mode collapse and training instability.

In recent years, with the rapid development of neural networks~\cite{SD,li2024exploring,yang2023using,dong2023entity,zhang2021lightweight}, Diffusion-based methods make significant strides in tasks such as image, video, and motion generation~\cite{ho2022imagen,ma2023follow,ma2022visual,ma2023magicstick,he2024strategic,he2023reti,SunN0JWN23,xu2023chain}.
FineDance\cite{finedance} and EDGE\cite{edge}  introduce Diffusion to generate diverse and high-quality dance clips of seconds, but they only focus on local motion quality of detailed dance clips and cannot quickly generate long-term dance movements that conform to the overall choreography rules.


\section{Method}
\label{sec:method}

\subsection{Preliminaries}

\begin{figure*}[t]
    \centering
    \includegraphics[width=\textwidth]{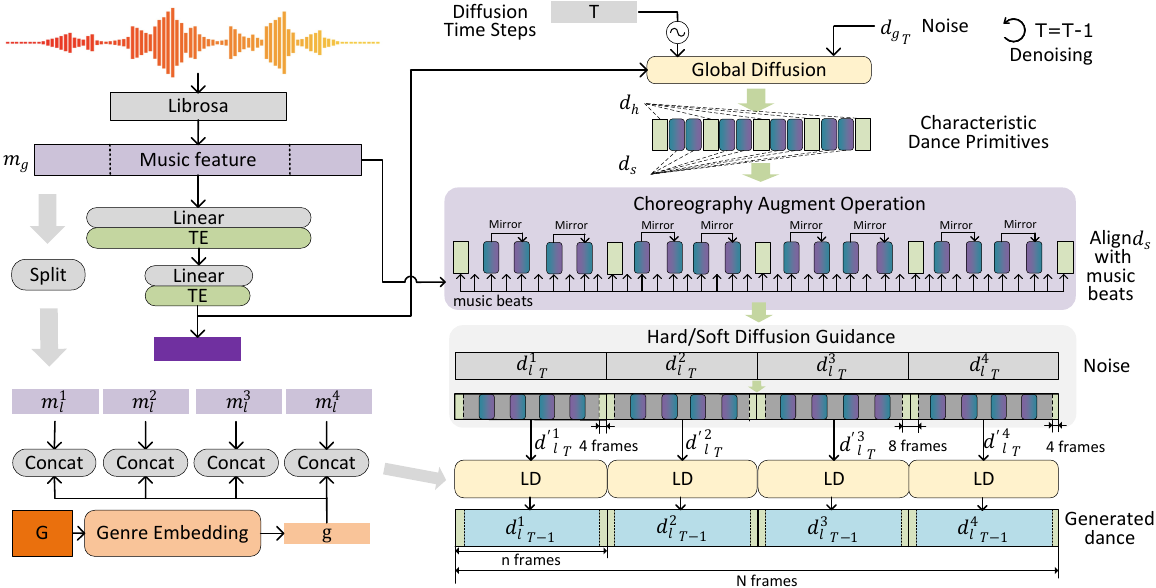}
    \caption{An overview of our framework. ``TE" is Transformer Encoder, ``G" is the genre of dance, ``LD" is the Local Diffusion Model.}
    \label{fig:Method}
\end{figure*}

\noindent\textbf{Music and Dance Representation.} Given a music clip, we follow~\cite{aist++} and employ Librosa \cite{librosa} to extract the music 2D feature map $\bm{m} \in \mathbb{R}^{ L \times 35}$, in which $L$ is the frame number and $35$ is the music feature channels with 1-dim envelope, 20-dim MFCC,  12-dim chroma, 1-dim one-hot peaks, and 1-dim one-hot beats. 
In addition, we follow EDGE~\cite{edge} and represent dance as $\bm{d} \in \mathbb{R}^{L\times 139}$. This motion representation obeys the SMPL\cite{SMPL} format (without fingers) and consists of the following components: (1) 4-dim foot-ground contact binary label, corresponding to left toe, left heel, right toe, right heel, where 1 means contact with ground and 0 means no contact; (2) 3-dim root translation; (3) 132-dim rotation information in 6-dim rotation repersentation~\cite{6drot}, the first 6-dim is global rotation and the remaining 126 dimensions correspond to the relative rotations of 21 sub-joints propagated along the kinematic chain.

\noindent\textbf{The Diffusion Model.}
We follow DDPM~\cite{DDPM} and EDGE~\cite{edge} to build our dance generation model. 
The diffusion model consists of two main processes: a diffusion process and a denoising process. 
The diffusion process perturbs the ground truth dance data  $\bm{d}_0$ into $\bm{d}_t$ over $t$ steps, we follow \cite{DDPM} to simplify this multi-step diffusion process into one step, which can be formulated as:
\begin{equation}
  \begin{split}
            \begin{array}{lr}
             q\left(\bm{d}_{t}| \bm{d}_{0}\right) =\mathcal{N}(\sqrt{\bar{\alpha}_{t}}\bm{d}_{0},{(1-\bar{\alpha}_{t})} \bm{I}),
            \end{array}
  \end{split}
\end{equation}
where $\bar{\alpha}_{t}$ is within the range of $(0,1)$ and follows a monotonically decreasing schedule. 
$\bar{\alpha}_{t}$ converges to $0$ as $t$ goes to infinity, making $\bm{d}_{t}$ converging to a sample from the standard normal distribution.
The denoising process employs a Transformer base-network  $f_\theta$ to gradually recover the motion, generating ${\hat{\bm{d}}}_0$ conditioned on given music $\bm{m}$. 
Instead of predicting the noise \cite{motiondiffuse}, we directly predict the ${\hat{\bm{d}}}_0$ like \cite{edge}. Therefore, the training process can be formulated as:

\begin{equation}
    \mathcal{L}_{\text{recon}}=\mathrm{E}_{\bm{d}_0, t}[\left\|{\bm{d}_0}-{f_\theta}\left(\bm{d}_t, t, \bm{m}\right)\right\|_2^2].
\end{equation}

\noindent\textbf{Choreography Rules.} Based on suggestions from professional choreographers and existing literature\cite{dance_design,dance_design2,blom1982intimate,choreomaster}, we want to generate long-duration dances that obeyed these three basic choreographic rules:
(1) The overall genre of the music and the dance should be consistent, conveying similar moods and tones.
(2) The beat of the music and the dance should be the same as far as possible. 
(3) The arrangement of dance should align with the structure of the accompanying music. For instance, identical meters in a musical phrase often correspond to symmetrical movements.

\subsection{Two-stage Dance Generation}
Given a extremely long music feature $\bm{m} \in \mathbb{R}^{L\times 35}, L=kN$, we first split $\bm{m}$ into segments of length $N$ without overlaps, i.e. $\left\{{{\bm{m}}_g^i \in \mathbb{R}^{N\times35}}\right\}_{i=1}^{k}$.
Our goal is to learn a neural network Loge, ${\bm{d}}_g^i=Lodge({m}_g^i), {\bm{d}}_g\in \mathbb{R}^{N\times 139}$, ${\bm{d}}=concatenate( \lceil{{\bm{m}}_g^i }\rceil,dim=0)$,  which means Lodge can parallelly generate extremely long dance sequences ${\bm{d}} \in \mathbb{R}^{kN\times 139}$ with a single inference.

\noindent\textbf{Method Overview.} 
In order to simultaneously consider both the global choreographic rules and the local dance details, we design a coarse to fine diffusion network with two stages as shown in Figure \ref{fig:Method}. The first stage is the global diffusion, which uses the global music feature ${\bm{m}}_g$ to learn the choreography patterns and produce characteristic dance primitives. The dance primitives are expressive key motions ${\bm{m}}_k\in \mathbb{R}^{8\times 139}$ with a higher motion kinematic energy, where 8 is the frame number.
Then, we perform choreographic augment operations on these dance primitives by the following three steps:
(1) We categorize them into hard-cue key motions $d_h$ that support parallel generation and soft-cue key motions $d_s$ that enhance the dance performance. (2) Based on the second choreographic rule, we mirror these soft-cue key motions.
(3) Based on the third choreographic rule,  we align soft-cue key motions to  the timing of the musical beats.

The second stage is the Local Diffusion (LD), which focuses on the quality of short-duration $n$ frames dance generation, corresponding to  several seconds. We further split each $\bm{m}_g$ into $\left\{{\bm{m}_l^j\in\mathbb{R}^{n\times35}}\right\}_{j=1}^{\lceil N/n \rceil}$.
As shown in Figure \ref{fig:Method}, we use the characteristic dance primitives as an intermediate-level representation of our two-stage diffusion network.
During the inference process, we replace the movements at the beginning and end of $d_t$, as well as those at the timing of musical beats, with these dance primitives. 
This way, we transfer globally learned choreographic patterns and expressive dance primitives obtained by global diffusion to local diffusion in a diffusion guidance manner.
Specifically, the hard-cue key motion uses the diffusion inpainting technique~\cite{repaint, edge} to control the start and end movements of the local diffusion. 
Meanwhile, during the diffusion denoising process, soft-cue key motions only serve as guidance in the initial $1000 \times s $ steps, where 1000 is the diffusion denoising steps. By adjusting the hyperparameter `s', we can control the extent to which local diffusion is influenced by these soft-cue key motions.
Notably, thanks to the hard cue motions, we can parallelly generate dance sequences $d$ much longer than N with a single inference.

\subsection{Global Diffusion}

\begin{figure*}[t]
    \centering
    \includegraphics[width=\textwidth]{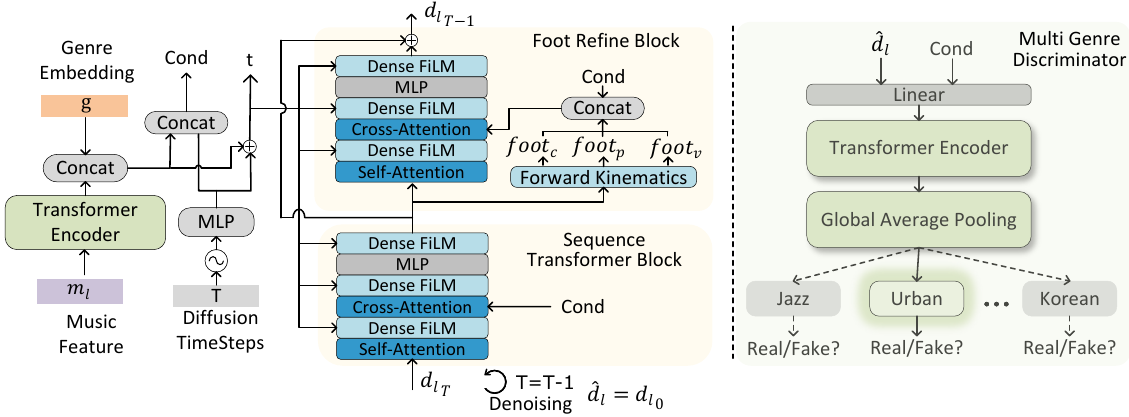}
    \caption{Training process of Local Diffusion.}
    \label{fig:LocalDiffusion}
\end{figure*}

Previous works overlook the global dependencies between music and dance, focusing only on the music-dance relationship within a small window. 
To address this issue, we only task global diffusion to generate sparse dance primitives. Subsequently, multiple local diffusions work in parallel to generate complete long dances.

Given a global music features ${\bm{m}}_g$ extracted by the Librosa\cite{librosa}. We feed ${\bm{m}}_g$ into a Transformer downsample network, which comprises a Linear layer and a Transformer Encoder Layer. Next, the compressed global music feature is sent to global diffusion. We adopte the EDGE framework as the foundation for global diffusion, making a single modification by adjusting the training objective to output sparse dance primitives. These primitives are key motions with only 8 frames, categorized as ${\bm{d}}_h$ and ${\bm{d}}_s$.  

There are key motions and transition motions in dance, where key motions are those with velocity curves near local minima, displaying greater expressiveness and richer semantic information, while transition motions are relatively monotonic.
To ensure that global diffusion concentrates solely on generating expressive key motions. 
We separated the expressive key movements and monotonous transitional movements in the dataset and trained global diffusion with only the expressive key motions.
Since the global diffusion learns key motions on a global scale, it already implicitly captures some choreographic patterns. To further enhance the overall dance coherence, we do choreography augment operation on ${\bm{d}}_s$, guiding the local diffusion to produce dance that more closely adheres  to choreography rules.


\subsection{Local Diffusion}
\noindent\textbf{Training Process.}
Thanks to our coarse to fine diffusion architecture, the local diffusion only needs to train the network on $n$ frames, which greatly accelerates the training speed and allows local diffusion to focus on the details of the dance movements for a few seconds.
The training process of local diffusion can be seen in Figure \ref{fig:LocalDiffusion}. We follow EDGE to build the Sequence Transformer Block, which consists  of self-attention layer\cite{transformer}, cross-attention layer\cite{saharia2022photorealistic}, multi-layer perception layer and the feature-wise linear modulation (FilM)\cite{film}.

In addition to the reconstruction loss, we introduce several other losses to enhance training stability and physical realism like previous works \cite{MDM,edge}. We compute the positional coordinates ${\bm{d}}_{\text{joint}}^{(i)}$ of the human body joints using forward kinematics, and then get the joint velocity ${\bm{d}}_{\text{j-vel}}^{(i)}$ and joint acceleration ${\bm{d}}_{\text{j-acc}}^{(i)}$. We then add the following loss functions: joint position Eq. (\ref{eq:jointsloss}), velocity Eq. (\ref{eq:vel}), and acceleration Eq. (\ref{eq:acc}):

\vspace{-1.5mm}
\begin{equation}
\begin{aligned}
{\bm{d}}_{\text{joint}}^{(i)} &= FK(\bm{d}^{(i)} ), 
\end{aligned}
\label{eq:joints}
\vspace{-1.5mm}
\end{equation}

\vspace{-1.5mm}
\begin{equation}
\begin{aligned}
\mathcal{L}_{\text{joint}}=\frac{1}{n} \sum_{i=1}^{n} \left\| {\bm{d}}_{\text{joint}}^{(i)} - {\hat{\bm{d}}}_{\text{joint}}^{(i)} \right\|^2_2,
\end{aligned}
\label{eq:jointsloss}
\vspace{-1.5mm}
\end{equation}

\begin{equation}
\begin{aligned}
\mathcal{L}_{\text{j-vel}}=\frac{1}{n-1} \sum_{i=1}^{n-1} \left\| {\bm{d}}_{\text{j-vel}}^{(i)} - {\hat{\bm{d}}}_{\text{j-vel}}^{(i)} \right\|^2_2,
\end{aligned}
\label{eq:vel}
\vspace{-1.5mm}
\end{equation}

\begin{equation}
\begin{aligned}
\mathcal{L}_{\text{j-acc}}=\frac{1}{n-2} \sum_{i=1}^{n-2} \left\| {\bm{d}}_{\text{j-acc}}^{(i)} - {\hat{\bm{d}}}_{\text{j-acc}}^{(i)} \right\|^2_2.
\end{aligned}
\label{eq:acc}
\vspace{-1.5mm}
\end{equation}

To optimize the contact between feet and the ground, we follow LEMO\cite{lemo, xiang} in decoupling the horizontal and vertical velocities of the feet, and optimizing the horizontal velocity $f_{hv}$ and downward vertical velocity $f_{dv}$ to 0 when the feet contact with the ground. 
\vspace{-1.5mm}
\begin{equation}
\begin{aligned}
\mathcal{L}_{\text{contact}}=\frac{1}{n-1} \sum_{i=1}^{n-1} \left\|  ({\hat{{\bm{f}}}}_{{hv}}^{(i)} + {\hat{{\bm{f}}}}_{{dv}}^{(i)})  \cdot {\hat{b}^{(i)}} \right\|^2_2,
\end{aligned}
\label{eq:contact}
\vspace{-1.5mm}
\end{equation}
where ${\hat{b}}$ is the predicted foot contact label.
Our overall training object is the weighted sum of the losses:
\vspace{-1.5mm}
\begin{equation}
\begin{aligned}
\mathcal{L}_{\text{total}} =&\mathcal{L}_{\text{recon}}  
+ {\lambda}_{\text{joint}} \mathcal{L}_{\text{joint}} 
+ {\lambda}_{\text{j-vel}} \mathcal{L}_{\text{j-vel}} \\
&+{\lambda}_{\text{j-acc}} \mathcal{L}_{\text{j-acc}} 
+ {\lambda}_{\text{contact}} \mathcal{L}_{\text{j-contact}}
+ {\lambda}_{\text{genre}} \mathcal{L}_{\text{genre}},
\end{aligned}
\label{eq:total}
\vspace{-1.5mm}
\end{equation}
where ${\lambda}_{\text{genre}}$ is formulated as Eq. (\ref{eq:genre}).

\noindent\textbf{Foot Refine Block.} The motion is expressed in the SMPL format, facilitating driven various human models and rendering. 
However, representing motion in the SMPL format involves a sequence of relative rotations and motion tree propagation. Small rotations near the root nodes, such as the legs and knees, result in significant rotations at the feet. Especially in dance movements, which involve a variety of foot actions, these challenges make it difficult for us to straightforwardly resolve foot skating issues by simply using foot-related loss functions.
We argue the main issue lies in the domain gap between the optimization objective and the data representation. The contact status between the feet and the ground is measured in a linear space based on joint positions, while the motion in the SMPL format exists in a nonlinear rotation space.
To tackle this, we introduce the Foot Refine Block inspired by \cite{xiang}. This module first computes the positions of foot keypoints $\bm{foot}_p$ through forward kinematics, as well as  foot velocity $\bm{foot}_v$. Then we calculate the foot-ground contact score $\bm{foot}_c$ follow \cite{xiang}. Building upon this, the Cross Attention mechanism is employed to further optimize foot movements. 

\noindent\textbf{Multi Genre Discriminator.} 
Local diffusion can produce high-quality, diverse dance segments.
As shown in the Figure \ref{fig:LocalDiffusion},
to ensure consistency with the overall musical style, we also concatenate the genre embedding $g$ with the music features, resulting in ${\bm{m}}_l^g$ as the condition for local diffusion. We then use a multi-genre discriminator (MGD) to control the dance genre following  MNET\cite{kim2022brand}. The training process of MGD can be formulated as:
\vspace{-1.5mm}
\begin{equation}
\begin{aligned}
&\mathcal{L}_{\text {genre }}=  \mathbb{E}_{\hat{\textbf{\textit{d}}_l}}
\left[\log MGD\left({\hat{\textbf{\textit{d}}_l}}, g,{\textbf{\textit{m}}_l}\right)\right]+ \\
&\mathbb{E}_{{\hat{\textbf{\textit{d}}_l}} ,  t}\left[\log \left(1-     
MGD\left(LD\left({\hat{\textbf{\textit{d}}_l}}, g, {\textbf{\textit{m}}_l}\right), g, {\textbf{\textit{m}}_l}\right)\right)\right],
\end{aligned}
\label{eq:genre}
\vspace{-1.5mm}
\end{equation}

\noindent\textbf{Parallel Inference.}
Given the input ${m}_l^j, g$ and corresponding ${d}_h, {d}_s$, the local diffusion outputs ${d}_l^j$. By concatenating $\left\{{{m}_l^j}\right\}_{j=1}^{N/n}$  along the time dimension, we obtain $d_g$. 
For simplicity in description, we omit `$j$' in subsequent writing.
To achieve parallel generation of long dance sequences, we divide $d_h$ into the first four frames and the last four frames. The first four frames serve as the tail four frames of the previous $d_l$, and the last four frames of $d_s$ serve as the leading four frames for the next $d_l$. This approach requires the local diffusion to generate the intervening dance motions coherently. However, directly using Diffusion inpainting techniques to control the first and last frames of each segment results in incoherent motions. To address this issue, we use a joint acceleration loss $\mathcal{L}_{\text{j-acc}}$ and incorporate a fine-tuning stage.
In this stage, we mixture $\bm{d_t}$ of Local Diffusion and
the ground truth $\bm{{d_l}_0}$ by
$\bm{{d_l}^\prime}_t [:4] = {\bm{{d_l}}_0}[:4]$, 
$\bm{{d_l}^\prime}_t [-4:] = {\bm{{d_l}}_0}[-4:]$,
$\bm{{d_l}^\prime}_t [4:-4] = {\bm{{d_l}}_t}[4:-4]$.
The $\mathcal{L}_{\text{recon}}$ loss in the fine-tuning stage is formulated as:
\vspace{-2mm}
\begin{equation}
\begin{aligned}
    \mathcal{L}_{\text{recon}} = \mathrm{E}_{\bm{{d_l}}_0, t}[\left\|{\bm{{d_l}}_0} - {f_\theta}\left(\bm{{d_l}^\prime}_t, g,t, \bm{m_l}\right)\right\|_2^2].
\vspace{-3mm}
\end{aligned}
\end{equation}
\section{Experiment}
\label{sec:experiment}


\subsection{Experimental Setup}
\noindent\textbf{Datasets.} We validate our method using the public music-dance paired dataset FineDance\cite{finedance} and AIST++\cite{aist++}. 
FineDance employs professional dancers to perform the dances and capture the data with an optical motion capture system.
The currently available dance data of FineDance contains 7.7 hours, totaling 831,600 frames, with a frame rate of 30 fps, and includes 16 different dance genres. The average dance length of FineDance is 152.3 seconds compared to 13.3 seconds for the AIST++ dataset, so we use the FineDance dataset to train and test the long-term dance generation algorithm.
We test the 20 pieces of music in the test set of the FineDance dataset and generate dance sequences with a length of 1024 frames (34.13 seconds). 

AIST++ is also a widely used dance dataset, containing 5.2 hours of dance data, with a frame rate of 60 fps, and includes 10 dance genres.


\noindent\textbf{Implementation details.} 
In the experiments on the \textbf{FineDance} dataset, the global music feature length $N$ is 1024, corresponding to 34.13 seconds; the local music feature length $n$ is 256, corresponding to 8.53 seconds. The global diffusion output 13 characteristic dance primitives, where 5 are $d_h$ and 8 are $d_s$. After the choreography augments operation, $d_s$ is mirrored to produce 16 instances, and it is aligned with the music's beat. 
The optimizer of global diffusion and local diffusion are Adan\cite{adan}, we use the  Exponential Moving Average(EMA) \cite{ema} strategy to make the loss convergence process more stable. The learning rate is $1e-4$. In the inference phase, we have two diffusion sampling strategies DDPM \cite{DDPM} and DDIM \cite{DDIM} that can be used to generate dance.
On the \textbf{AIST++} dataset, we downsampled the dance to 30 fps for training. Then we generated dances with 30 fps. Finally, we interpolated the output dances to 60 fps and followed the experimental setup of Bailando~\cite{bailando} for testing.
The music-dance data from AIST++ has been segmented into numerous short clips. 
Therefore, we change the global music feature length $N$ to be 256 and the global music feature length $n$ to be 128.

\subsection{Comparisons on the FineDance dataset}

As shown in Table \ref{tab:results}, we compare our method with the advanced existing works. FACT \cite{aist++} and MNET \cite{kim2022brand} are auto-gressive dance generation methods. Bailando \cite{bailando} is an outstanding music-driven dance generation algorithm. It employs VQ-VAE to transform dance movements into tokens. Subsequently, a GPT model forecasts this token sequence, which is then decoded to render the final dance.
To the best of our knowledge, EDGE \cite{edge} is a diffusion-based dance generation algorithm, achieving the strongest qualitative performance in short-duration dance generation. 
During the diffusion denoising process, they assign the latter half of the previous dance segment to the first half of the subsequent segment, and utilize interpolation to maintain consistency, thereby achieving long-term dance generation. 

\begin{table*}
\resizebox{\linewidth}{!}{
\begin{tabular}{lccccccccc}
    \toprule[1.5pt]
    \noalign{\smallskip}
    \multirow{2}*{Method} & \multicolumn{3}{c}{Motion Quality} & \multicolumn{2}{c}{Motion Diversity} & \multirow{2}*{BAS$\uparrow$} & \multirow{2}*{Run Time$\downarrow$} & \multirow{2}*{Wins$\uparrow$} \\ \cmidrule(lr){2-4} \cmidrule(lr){5-6}
    & $\mathrm{FID}_k\downarrow$ & $\mathrm{FID}_g\downarrow$ & Foot Skating Ratio$\downarrow$ & $\mathrm{Div}_k\uparrow$ & $\mathrm{Div}_g\uparrow$ & & & \\
    \noalign{\smallskip}\midrule
    \noalign{\smallskip}
    Ground Truth & / & / & 6.22$\%$ & 9.73 & 7.44 & 0.2120 & / & 42.6$\%$ \\
    \midrule
    FACT\cite{aist++} & 113.38 & 97.05 & 28.44$\%$ & 3.36 & 6.37 & 0.1831 & 35.88s & 96.7$\%$ \\
    MNET\cite{kim2022brand} & 104.71 & 90.31 & 39.36$\%$ & 3.12 & 6.14 & 0.1864 & 38.91s & 92.3$\%$ \\
    Bailando\cite{bailando} & 82.81 & \textbf{28.17} & 18.76$\%$ & 7.74 & 6.25 & 0.2029 & 5.46s & 68.2$\%$ \\
    EDGE\cite{edge} & 94.34 & 50.38 & 20.04$\%$ & \textbf{8.13} & 6.45 & 0.2116 & 8.59s & 80.6$\%$ \\
    \midrule
    \textbf{Lodge} (DDIM) & 50.00 & 35.52 & \textbf{2.76$\%$} & 5.67 & 4.96 & 0.2269 & \textbf{4.57}s & / \\
    \textbf{Lodge} (DDPM) & \textbf{45.56} & 34.29 & {5.01$\%$} & 6.75 & 5.64 & \textbf{0.2397} & 30.93s & / \\
    \bottomrule[1.5pt]
\end{tabular}
}
\caption{Compare with SOTAs on the FineDance dataset. Wins is the ratio of victories Lodge(DDPM) achieved in the user study.}
\label{tab:results}
\end{table*}

\noindent\textbf{Motion Quality.} 
To evaluate the motion quality of generated dance sequences, we follow the previous methods\cite{aist++, bailando} to calculate the Frechet Inception Distance (\textbf{FID})\cite{fid} distance between motion features of the generated dance and the ground truth dance sequences.
The previous methods such as \cite{bailando} calculate kinetic\cite{kinfeature} and geometric\cite{geofeature}  motion features using the global coordinates of all the SMPL\cite{SMPL} joints, which is suitable for measuring the quality of movements lasting only a few seconds. However, for longer motion sequences, where trajectories become more complex, this measurement approach focuses too heavily on root positions, neglecting local movements and resulting in data that lacks comparability. 
Therefore, we use the global coordinates of the root joint and the relative distances of other child joints to compute kinetic and geometric features.
The kinematic feature (subscript `k'), indicates the speed and acceleration of the movement and reflects the physical characteristics of the dance. Therefore the FID distance between kinematic features \textbf{$\mathrm{FID}_k$} measures the physical reality of the motion.
The geometric feature (subscript `g'), is calculated based on multiple predefined movement templates, thus the FID distance between geometric features \textbf{$\mathrm{FID}_g$}  reflects the quality of the overall dance choreography.
In addition, we follow \cite{GMD} to report the \textbf{Foot Skating Ratio (FSR)}, which measures the proportion of frames in which either foot skids more than a certain distance while maintaining contact with the ground (foot height $<$ 5 cm).


\noindent\textbf{Motion Diversity.} 
To evaluate the motion diversity of generated dance sequences, we calculate the mean Euclidean distance within the motion feature space, as outlined in the works of Bailando\cite{bailando}. 
$\mathrm{DIV}_k$ represents the motion diversity in the kinematic feature space, while $\mathrm{DIV}_g$ denotes the diversity in the geometric feature space.
Table \ref{tab:results} reveals that our Lodge approach achieved the highest $\mathrm{DIV}_g$ score, which can be credited to our adoption of global diffusion and characteristic dance primitives for mastering diverse choreography patterns.

\noindent\textbf{Beat Alignment Score (BAS).} 
To evaluate the beat consistency between the generated dance and the given music, we follow \cite{aist++} and use the BAS to evaluate our methods, our approach demonstrated the highest Beat Alignment Score of 0.2397.

\noindent\textbf{Production efficiency.}
In our inference process, we evaluated the average \textbf{Run time} taken for model generation. To ensure fairness in testing, we excluded data preprocessing time from our calculations. All experiments were conducted on the same computer equipped with an Nvidia A100 GPU and 256GB of memory.

Run Time in Table \ref{tab:results} presents the average Run Time required to generate 1024 frames of dance movements. Bailando achieved a outstanding performance, but its runtime increases linearly with the length of the sequence generated. EDGE, using the DDIM accelerated sampling strategy and linear interpolation, also achieved a fast level for generating long dance sequences.
Our method uses DDPM sampling with a denoising step of 1000, taking 30.93 seconds. Using DDIM with 50 denoising steps takes only 4.57 seconds. Meanwhile, our parallel architecture ensures runtime remains stable even with longer sequences.

\noindent\textbf{User study.}
We conducted a user study where 20 participants viewed 17 video pairs.  Each pair consists of two dance sequences: one created by Lodge (DDPM) and the other by different methods or ground truth. 

\subsection{Comparisons on the AIST++ dataset}
As Table ~\ref{tab:aistcompare} shows, we train Lodge on AIST++ and compare it with SOTAs.
Due to the lack of long-duration dance in the AIST++ dataset, Lodge's performance does not reach the best metrics.
However, compared to our baseline model EDGE, Lodge shows improvement in multiple metrics.

\begin{table}
\resizebox{0.99\linewidth}{!}{
\begin{tabular}{lcccccc}
    \toprule[1.5pt]
    \noalign{\smallskip}
    \multirow{2}*{Method} & \multicolumn{2}{c}{Motion Quality} & \multicolumn{2}{c}{Motion Diversity} & \multirow{2}*{BAS$\uparrow$} & \\ \cmidrule(lr){2-3} \cmidrule(lr){4-5}
    & $\mathrm{FID}_k\downarrow$ & $\mathrm{FID}_g\downarrow$ & $\mathrm{Div}_k\uparrow$ & $\mathrm{Div}_g\uparrow$ & &\\
    \noalign{\smallskip}\midrule
    \noalign{\smallskip}
    Ground Truth & 17.10 & 10.60 & 8.19 & 7.45 & 0.2374 \\
    \midrule
    Li \etal~\cite{li2020learning} & 86.43 & 43.46 & 6.85 & 3.32 & 0.1607 \\
    DanceNet~\cite{zhuang2020music2dance} & 69.18 & 25.49  & 2.86 &  2.85 & 0.1430 \\
    DanceRevolution~\cite{huang2020dance} & 73.42 & 25.92 & 3.52 & 4.87 & 0.1950 \\
    FACT~\cite{aist++} & 35.35 & 22.11 & 5.94 & 6.18 & 0.2209 \\
    Bailando~\cite{bailando} & \textbf{28.16} & \textbf{9.62} & \textbf{7.83} & \textbf{6.34} & 0.2332 \\
    EDGE~\cite{edge} & 42.16 & 22.12 & 3.96 & 4.61 & 0.2334 \\
    \midrule
    \textbf{Lodge} (DDPM) & 37.09 & 18.79 & 5.58 & 4.85 & \textbf{0.2423} \\
    \bottomrule[1.5pt]
\end{tabular}
}
\vspace{-2mm}
\caption{Compare with SOTAs on the AIST++ dataset.}
\label{tab:aistcompare}   
\end{table}

\subsection{Ablation Studies}
In this section, we use DDPM sampling strategy and perform ablation experiments on the FineDance dataset to evaluate the different parts: (1) the characteristic dance primitives, (2)the soft cue guidance, (3) the foot refine block. 

\noindent\textbf{Effect of the characteristic dance primitives.}
We conducted a series of ablation experiments to validate the effect of the characteristic dance primitives. In Table \ref{tab:Primitives}, `C' indicates we use characteristic dance primitives to guide the local diffusion, `M' represents we mirror the characteristic dance primitives. `B' denotes beat alignment, we align the characterized dance primitives with the music's beats, guiding the Local diffusion to generate more expressive movements at these beat points. 
If beat alignment is not applied, then the characterized dance primitives are uniformly distributed across various timelines to guide the local diffusion.

The first row in Table \ref{tab:Primitives} shows the results when not using characteristic dance primitives, relying solely on some $d_h$ for parallel long action generation but not using $d_s$ within a Local diffusion. This scenario leads to lower quality of motion (FID), diversity, and music rhythm alignment metrics. In contrast, rows two and three, which incorporate guidance from characteristic dance primitives, display significant improvements in Div and BAS. This improvement is because characteristic dance primitives are expressive key motions; their inclusion helps prevent the neural network from generating average, monotonous movements. The last row, achieving the optimal results, demonstrates the effectiveness of our strategy.


\begin{table}[th]
\centering
\resizebox{0.33\textwidth}{!}{
	\begin{tabular}{lcccccccc}
	\toprule [1pt] 
	\multicolumn{3}{c}{Ablations} & \multicolumn{3}{c}{Metrics} \\ \cmidrule(lr){1-3}   \cmidrule(lr){4-6}   
        C  & M & B &   $\mathrm{FID}_k\downarrow$ &   $\mathrm{Div}_k\uparrow$  &  BAS $\uparrow$ \\
	\noalign{\smallskip}\hline\noalign{\smallskip}
         &  &   &60.91  &5.16  &0.2090\\
         \checkmark & \checkmark &   &60.20  &5.54  &0.2132\\
         \checkmark &  &  \checkmark &52.18  &5.75 &0.2139\\
		\checkmark & \checkmark & \checkmark &45.56  &6.75  &0.2397\\
		\bottomrule [1pt] 
               \noalign{\smallskip}
	\end{tabular}
	}
        \vspace{-2mm}
	\caption{Ablation study of the characteristic dance primitives.}
	\label{tab:Primitives}   
    \vspace{-3mm}
\end{table}


\noindent\textbf{Effect of the Soft-cue Guidance.}
Our soft cue guidance weight can be adjusted using the hyperparameter `s', where a larger `s' value signifies a stronger effect. Table \ref{tab:softguidance} demonstrates the outcomes resulting from setting various `s' values.
With the increase in `s', there is a corresponding enhancement in $\mathrm{FID}_k$ and Beat Alignment Score. The optimal performance is achieved when `s' is set to 1.

\begin{table}[th]
\centering
\resizebox{0.35\textwidth}{!}{
	\begin{tabular}{lcccccccc}
		\toprule [1pt] 
		Method &   $\mathrm{FID}_k\downarrow$ &   $\mathrm{Div}_k\uparrow$  &  BAS $\uparrow$ \\
		\noalign{\smallskip}\hline\noalign{\smallskip}
            Ground Truth  &/  &9.73& 0.2120\\
            \noalign{\smallskip}\hline\noalign{\smallskip}
            s=0    &60.91  &5.16  & 0.2090\\  
		s=0.05 &59.66  &5.43  & 0.2131\\
            s=0.25 &60.51  &5.41  & 0.2132\\
		s=0.5  &60.46  &5.35  & 0.2196\\
            s=0.75 &59.89  &5.32  & 0.2208\\
            s=0.95 &53.63  &5.37  & 0.2239\\
            s=1    &45.56  &6.75  & 0.2397\\
		\bottomrule [1pt] 
               \noalign{\smallskip}
	\end{tabular}
	}
        \vspace{-2.5mm}
		\caption{Ablation study of the soft cue guidance. }
	\label{tab:softguidance}   
        \vspace{-1mm}
\end{table}

\noindent\textbf{Effect of the Foot Refine Block.}
As shown in Table \ref{tab:footskating}, after incorporating the Foot Refine Block, 
the motion quality $\mathrm{FID}_k$ had a large improvement, especially the Foot Skating Ratio decreased from 5.94$\%$ to 5.01$\%$, which proves that our proposed Foot Refine Block can effectively improve the foot-ground contact quality and reduce the probability of foot skating phenomenon.

\begin{table}[th]
\vspace{-1mm}
\centering
\resizebox{0.45\textwidth}{!}{
	\begin{tabular}{lcccccccc}
		\toprule [1pt] 
		Method &   $\mathrm{FID}_k\downarrow$ &  $\mathrm{Div}_k\uparrow$  &  BAS $\uparrow$ &    Foot Skating Ratio$\downarrow$\\
		\noalign{\smallskip}\hline\noalign{\smallskip}
            Ground Truth  &/ &9.73 &0.2120 & 6.22$\%$\\
            \noalign{\smallskip}\hline\noalign{\smallskip}
            w/o Foot Refine Block &53.48 &6.20 & 0.2216 &5.94$\%$\\
            w. Foot Refine Block &45.56 &6.75 & 0.2397 &5.01$\%$\\
		\bottomrule [1pt] 
               \noalign{\smallskip}
	\end{tabular}
	}
        \vspace{-2mm}
		\caption{Ablation study of the foot refine block.}
	\label{tab:footskating}   
        \vspace{-2mm}
\end{table}

\section{Conclusion and Limitation}

In this work, we introduce Lodge, a two-stage coarse-to-fine diffusion network, and propose characteristic dance primitives as intermediate-level representations for the two diffusion models. 
Lodge has been extensively evaluated through user studies and standard metrics. 
Our generated samples demonstrate that Lodge can parallelly generate dances that conform to choreographic rules while preserving local details and physical realism. 
However, our method currently cannot generate dance movements with hand gestures or facial expressions, which are also crucial for performances. 
This limitation opens avenues for future research.

\section*{Acknowledgment}
This work was supported in part by the Shenzhen Key Laboratory of next generation interactive media innovative technology (No.ZDSYS20210623092001004), in part by the China Postdoctoral Science Foundation (No.2023M731957),  in part by the National Natural Science Foundation of China under Grant 62306165, in part by the the Peng Cheng Laboratory (PCL2023A10-2), in part by the NSFC project No.62125107.
{
    \small
    \bibliographystyle{ieeenat_fullname}
    \bibliography{main}
}
\clearpage
\appendix
\renewcommand\thesection{\Alph{section}}

\section{Details of the Training Process}
As shown in the figure below. We trained the two  stages separately to save graphics memory.
The Global Diffusion is trained on long music input and sparse key motions extracted from ground truth. The output key motions of Global Diffusion are categories in \bm{$d_h$} and \bm{$d_s$} to guide the Local Diffusion only in the inference phase.
\begin{figure}[!htbp]
  \centering
  \includegraphics[width=1.0\linewidth]{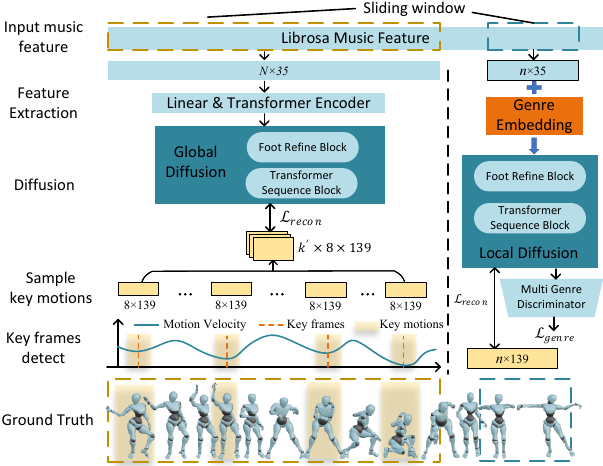}
  \caption{The Training process of Lodge.}
  \label{fig:training}
\end{figure}

\section{Details of the Hard/Soft Diffusion Guidance}
\label{sec:rationale}
We categorize the characteristic dance primitives generated by global diffusion into hard-cue key motions $d_h$ and soft-cue key motions $d_s$. We employ distinct diffusion guidance strategies for each, enabling them to guide local diffusion.

The role of $d_h$ is to guide the local diffusion in generating the initial and final segments of the dance, ensuring that the concurrently generated dance fragments can seamlessly concatenate into a coherent, long-form dance. Therefore, we adopt Hard Diffusion Guidance for this purpose.

On the other hand, $d_s$ serves to provide guidance to local diffusion. In this case, we aim for the guidance to be flexible, avoiding any disruption to the coherence of the dance generated by local diffusion. Consequently, we propose the Soft Diffusion Guidance algorithm for $d_s$. As illustrated in the pseudocode below, our proposed soft diffusion operates only for the first $1000 \times (1-s)$ steps, where $s$ is a  hyperparameter. The impact of different $s$ values on the results is detailed in Table 3 of the main paper.

\begin{lstlisting}[language=Python, caption=Pseudocode of the Hard/Soft Diffusion Guidance]
import torch, librosa
# m is the given music feature, m.shape = [L, 35], L is the time length
m = m[:ln] # l = L//n, n is the output frame number of one local diffusion
d_h, d_s = GlobalDiffusion(m)
# d_h.shape = [(l+1),8,139]; d_s.shape = [2l,8,139]
d_h = d_h.reshape([(l+1)*8,139])
d_h = d_h[4:-4].reshape([l,8,139])
d_s = Mirror(d_s).reshape(4l,8,139)
# Get music beat index by the librosa toolkit
beats = librosa.beatidx(m)
value,mask = torch.zeros([l,n,139])
value[:,:4,:] = d_h[:,:4,:]
value[:,-4:,:] = d_h[:,-4:,:]
value[:, beats-4:beats+4,:] = d_s   
mask[:,:4,:] = 1   
mask[:,-4:,:] = 1   
mask[:, beats-4:beats+4,:] = 1  
def guidance_sample(m,value,mask,s):
    d = torch.rand([l,n,139])
    # There are 1000 diffusion steps.
    for i in reversed(range(0, 1000)):
    if i > 1000*(1-s):
        #  sample d from step t to step t-1
        d = p_sample(d, m, t)
        # The soft-cue diffusion guidance
        value_ = q_sample(value, t - 1)
        d = value_*mask+(1.0 - mask) * d
        # The hard-cue diffusion guidance
        d[:,:4] = value[:,:4]*mask[:,:4]+(1.0 - mask[:,:4] )*d[:,:4] 
        d[:,-4:] = value[:,-4:] *mask[:,-4:]+(1.0-mask[:,-4:])*d[:,-4:]
    else:
        d = p_sample(d, m, t)
        d[:,:4] = value[:,:4]*mask[:,:4]+(1.0-mask[:,:4])*d[:,:4] 
        d[:,-4:] = value[:,-4:]*mask[:,-4:]+(1.0-mask[:,-4:])*d[:,-4:]

    d = d.reshape([ln, 139])
    return d
\end{lstlisting}

\begin{figure}[!htbp]
  \centering
  \includegraphics[width=1.0\linewidth]{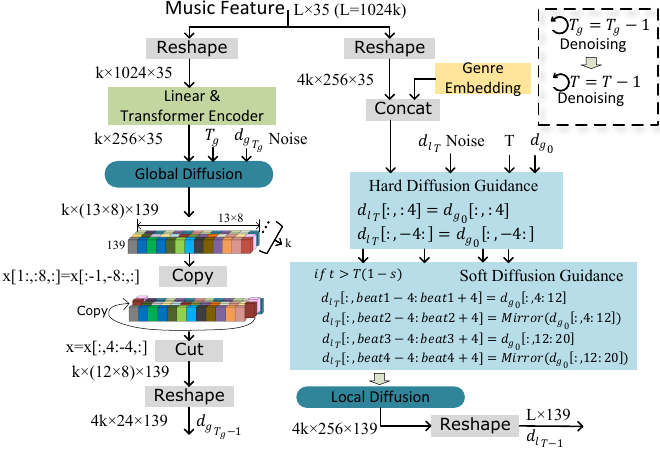}
  \caption{The inference process of Lodge.}
  \label{fig:inference}
\end{figure}

\section{Details of $d_s$ and $d_h$} 
Their primary distinction lies in different purposes.
The soft-cue key motion use \bm{$d_s$} to guide Local Diffusion to follow the overall choreographic patterns and increase motion expressiveness.
While the primarily purpose of hard-cue key motion \bm{$d_h$} is to support parallel generation.
Both \bm{$d_s$} and \bm{$d_h$} are 8-frame key motions generated by Global Diffusion. \bm{$d_h$} operates at the beginning and end of Local Diffusion, employing hard diffusion guidance to ensure strict consistency with the initial and final frames of the generated motion, thereby supporting parallel generation. Meanwhile, \bm{$d_s$}  operates in the middle of Local Diffusion, serving as a soft cue to improve the dance quality.


\section{Additional Ablation Studies (tested on the FineDance dataset)}

\subsection{The Characteristic Dance Primitives}
To reduce the computational load of Global Diffusion and to convey global choreography patterns effectively, we propose the Characteristic Dance Primitives. These primitives are dimensionalized as $(l^\prime, 8, 139)$, where $l^\prime$ represents the number of dance primitives, `8' denotes the temporal dimension encompassing a continuous sequence of eight frames, and `139' corresponds to the dimensions of the motion feature.
However, it is feasible to configure Dance Primitives as discrete frames.
Therefore, we conducted a four-fold temporal downsampling of the ground truth dance, which is utilized to train the Global Diffusion for generating discrete dance primitives. To evaluate the relative efficacy of these methodologies, we conduct ablation experiments on the dance primitives as Table \ref{tab:primi_format}.

\begin{table}[th]
\centering
\resizebox{0.4\textwidth}{!}{
	\begin{tabular}{lcccccccc}
	\toprule [1pt] 
	Method &   $\mathrm{FID}_k\downarrow$ &  $\mathrm{Div}_k\uparrow$  &  BAS $\uparrow$ \\
	\noalign{\smallskip}\hline\noalign{\smallskip}
        Ground Truth  &/  &9.73 & 0.2120 \\
        \noalign{\smallskip}\hline\noalign{\smallskip}
        Discrete   &55.17 &5.44 & 0.1969 \\
        Continuous &45.56 &6.75 & 0.2397 \\
	\bottomrule [1pt] 
        \noalign{\smallskip}
	\end{tabular}
        }
	\caption{Ablation study of the characteristic dance primitives. `Discrete' means the dance is generated by the guidance of discrete dance primitives, `Continuous' means the dance is generated by the guidance of continuous dance primitives}
	\label{tab:primi_format}   
\end{table}
 
The generated motion guided by discrete dance primitives often results in incoherence, primarily due to the lack of velocity information. This issue is reflected in the increased values of the $\mathrm{FID}_k$\cite{aist++, bailando} as shown in Table \ref{tab:primi_format}. Furthermore, the guidance provided by these discrete dance primitives disrupts the beat consistency between music and dance, which consequently leads to a significant decline in the Beat Alignment Score (BAS)\cite{aist++}.

\subsection{Ablation Studies of the Hyper-parameter $N$ and $n$}
As described in Section 3.2 of the main paper, $N$ represents the temporal receptive field of the Global Diffusion. The length of global music feature input into Global Diffusion is $N$. Meanwhile, $n$ denotes the frame number of dance generated by the Local Diffusion.

In this part, we investigate the impact of different $N$ and $n$. Thanks to our parallel architecture, Lodge can directly generate dance with $ln$ frames, where $l$ is a positive integer. The primary objective of these ablation experiments is to explore how different values affect dance performance. 

\begin{table}[th]
\centering
\resizebox{0.4\textwidth}{!}{
	\begin{tabular}{ccccccccc}
	\toprule [1pt] 
	$N$ & $n$ &  $\mathrm{FID}_k\downarrow$ &  $\mathrm{Div}_k\uparrow$  &  BAS $\uparrow$ \\
        \noalign{\smallskip}\hline\noalign{\smallskip}
        1024 & 512  & 61.66  &\textbf{8.14}  & 0.1864  \\
        1024 & 256  & \textbf{45.56}  &6.75  & \textbf{0.2397}  \\
        1024 & 128  & 45.86  &5.54  & 0.2212  \\
        512  & 256  & 59.72  &5.30  & 0.2182  \\
        512  & 128  & 46.74  &5.76  & 0.2124  \\
	\bottomrule [1pt] 
        \noalign{\smallskip}
	\end{tabular}
        }
        \caption{Ablation study of the hyper-parameter $N$ and $n$.}
        \label{tab:Nn}   
\end{table}

As shown in Table \ref{tab:Nn}, when $n$ is 512, the quality of motion, as measured by $\text{FID}_k$, deteriorates significantly due to the network's limited capability in modeling long sequences. This also results in a substantial increase in the cost of training Local Diffusion.
Comparing cases where $n$ is 128 and 256, we observe only a marginal difference in $\text{FID}_k$. However, crucially, we find that maintaining coherence at this value requires frequent incorporation of $d_h$
within the Hard Diffusion Guidance. Such regular intervention tends to disrupt the overall dance structure. Therefore, we ultimately set $n$ as 256.

Comparing the second and fourth rows, it's evident that when $N$ is set to 1024, all metrics show improved performance. Additionally, a larger $N$ enables more comprehensive modeling of the global dependencies between music and dance. Therefore, we ultimately set $N$ as 1024.

\section{Visualization Results}
\begin{figure}[h]
    \centering
    \includegraphics[width=0.5\textwidth]{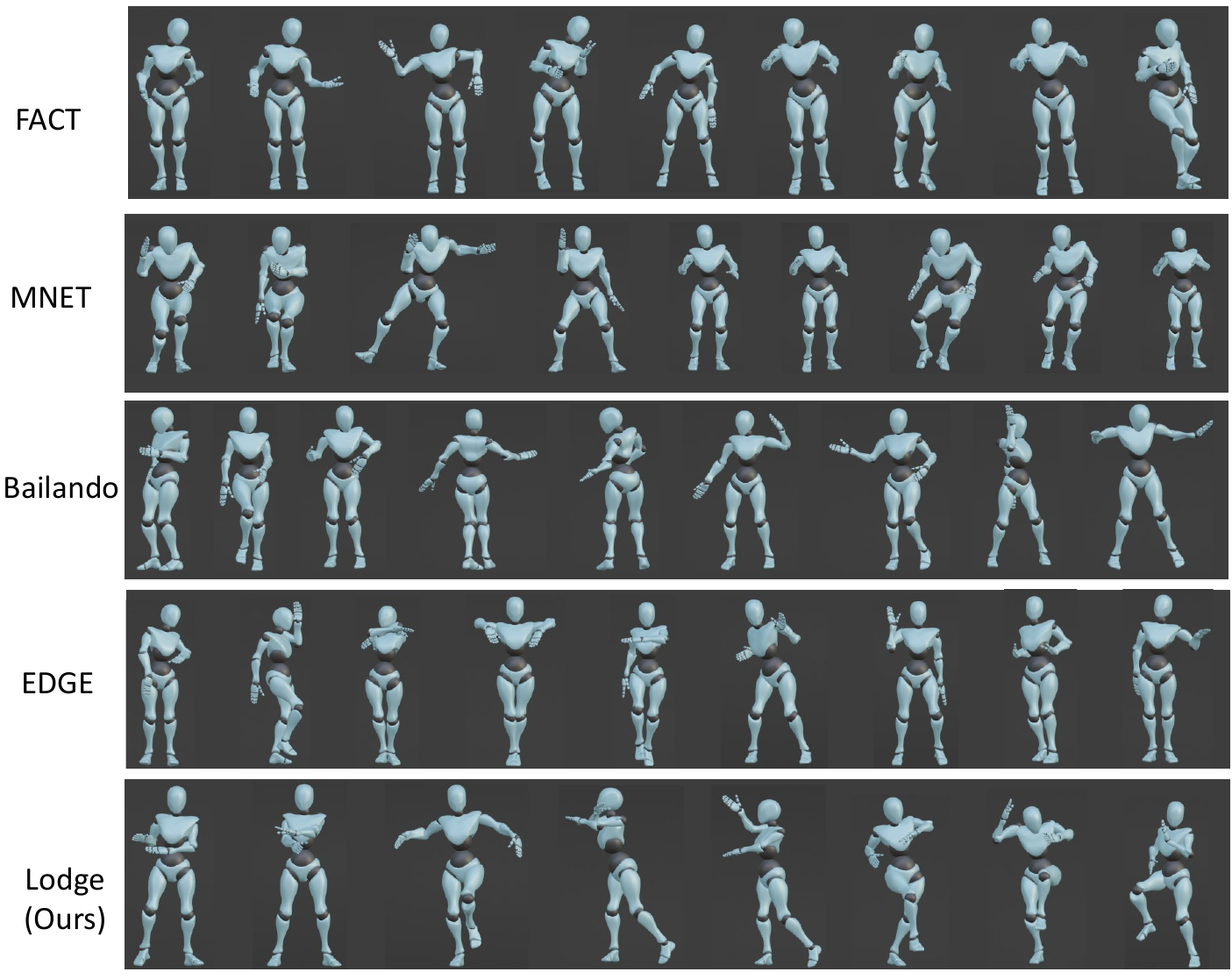}
    \caption{Compare with the SOTAs.}
    \label{fig:Datasets}
\end{figure}
We strongly wish you to watch the video in our project page for more details.
We conducted comparisons with state-of-the-art dance algorithms, including FACT\cite{aist++}, MNET\cite{kim2022brand}, Bailando\cite{bailando}, and EDGE\cite{edge}.
Both FACT and MNET are models based on the Transformer and autoregressive architecture. They encounter significant motion freezing issues during long-duration generation. After several seconds, their motion tends to freeze.
Bailando is a model designed based on VQ-VAE\cite{vqvae} and GPT\cite{gpt}. Its primary limitation lies in the encoding capacity of VQ-VAE, which restricts the network's ability to produce complex dance movements.
EDGE is a model based on Diffusion and serves as the backbone of this study. Its main issue is the lack of learning global choreography patterns, resulting in noticeable incoherence at the joints and a relative monotony in the movements.
Our method, benefiting from the Coarse-to-Fine architecture, along with the Characteristic Dance Primitives and the Foot Refine Block, is capable of generating coherent, high-quality, and expressive dance sequences.

\end{document}